\documentclass[]{interact}

\usepackage{graphicx} 
\usepackage{hyperref}   
\hypersetup{
	colorlinks=true,
	linkcolor=blue,
	citecolor=blue,
	urlcolor=blue,
}
\usepackage[square,numbers]{natbib}
\usepackage{epsfig}
\usepackage{setspace}
\usepackage{epstopdf}
\usepackage{epsfig}
\usepackage{amsfonts}
\usepackage{amsthm}
\usepackage{multirow}
\usepackage{rotating}
\usepackage{array}
\usepackage{longtable}

\usepackage{amssymb}

\newtheorem{MyDef}{Definition}
\newtheorem{MyTheo}{Theorem}
\newtheorem{MyLemma}{Lemma}
\newtheorem*{MyProof}{Proof}

\begin{document}

\title{The Effect of Sideslip on Jackknife Limits During Low Speed Trailer Operation}

\author{
  \name{Leng Zhe and Mark A. Minor\thanks{CONTACT M. Minor. Email:  mark.minor@utah.edu}}
    \affil{Robotic Systems Laboratory, Department of Mechanical Engineering, University of Utah, Salt Lake City, UT 84112}	
}

\maketitle    

\begin{abstract}
Jackknifing refers to the serious situation where a vehicle-trailer system enters a jackknife state and the vehicle and trailer eventually collide if trailer operation is not corrected. This paper considers low speed trailer maneuvering typical of trailer backing where jackknife state limits can vary due to sideslip caused by physical interaction between the vehicle, trailer, and environment. Analysis of a kinematic model considering sideslip at the vehicle and trailer wheels indicates that vehicle-trailer systems should be divided into three categories based on the ratio of hitch length and trailer tongue length, each with distinct behaviors. The Long Trailer category may have no jackknifing state while the other two categories always have states leading to jackknifing. It is found that jackknife limits, which are the boundaries between the jackknifing state and the recoverable regions, can be divided into safe and unsafe limits, the latter of which must be avoided. Simulations and physical experiments support these results and provide insight about the implications of vehicle and trailer states with slip that lead to jackknifing. Simulations also demonstrate the benefit of considering these new slip-based jackknife limits in trailer backing control.
\end{abstract}


\section{Introduction}
Trailer backing is widely used in agriculture, industry, and recreation activities. Automated trailer operation has potential to improve efficiency and convenience, but it is important for these systems to recognize limiting conditions. When backing a trailer, the vehicle and trailer system may enter a state called ``jackknife." In this state, hitch angle magnitude will keep increasing if the system continues to back, regardless of steering angles. Physical damage will result if trailer backing continues. As a result, engineers establish conservative hitch angle limits to prevent the system from jackknifing. Steering commands are either limited to prevent the system from passing these boundaries or the vehicle pulls forward so that hitch angle can be recovered into a non-jackknifing region, whereupon trailer backing resumes. Jackknifing and its recovery reduces mobility and efficiency of trailer backing. Sideslip caused by interaction between the vehicle, trailer, and terrain shifts the jackknife limits during operation, making jackknifing behavior uncertain. Hence there is a need to better understand how jackknife limits are affected by slip.  

Jackknifing is governed by vehicle and trailer kinematic parameters combined with a kinematic model that describes how hitch angle varies as a function of steering angle. Sideslip, which results from quasi-static forces acting on the system as the system interacts with terrain, perturbs the kinematic model of the system and changes jackknifing behavior. Jackknife limits can be better determined with an improved understanding of how wheel slip affects those limits. As this paper shows, sideslip can skew the jackknife regions and results in different types of jackknife limits requiring different recovery actions, which depend on the lengths of the trailer tongue and hitch. For safe and accurate jackknife characterization, sideslip, trailer size, and tongue length must be considered. 

\begin{figure}
\centerline{\epsfig{figure=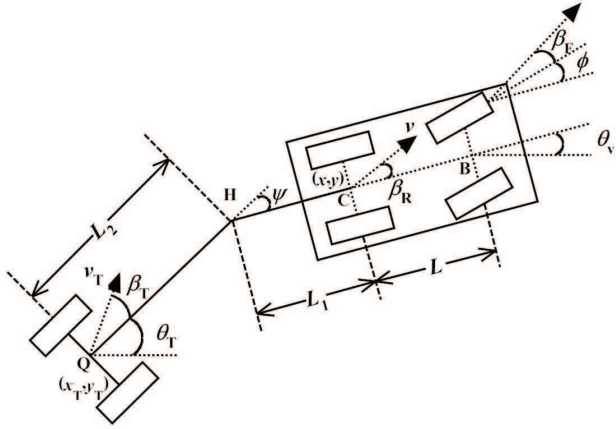,height=6cm}}
\vspace{-0.3 cm}
\caption{Kinematics of the vehicle-trailer system.}
\label{fig:model}
\end{figure}

In this paper, we analyze jackknife states in detail based on an extended kinematics model considering sideslip in the vehicle and trailer, Fig. \ref {fig:model}. The boundaries of the jackknife states, hereby called ``jackknife limits", are defined analytically. Vehicle and trailer systems are then divided into three categories based on hitch length and trailer tongue length. The jackknife limits for each category are derived individually with sideslip taken into consideration, indicating that there are actually safe and unsafe jackknife limits. The behaviors of the system when approaching and entering jackknife regions are then analyzed. Suggestions for dealing with the implications of these different regions and limits are then provided. Simulation results confirm the effect of sideslip on jackknife limits. Field experiments validate the predicted jackknife limits and illustrate usage of the analysis to estimate jackknife limits using sensor based sideslip estimates.  Derivations further highlight applicability of this research to both trailer backing and trailer pushing, where a trailer is manipulated from a hitch on the front of the vehicle.  

\subsection{Related Work}
As one of the practical challenges in autonomous trailer backing, jackknifing has been discussed in multiple works. Tanaka et al. \cite{Tanaka1998} and Matsushita and Murakami \cite{Matsushita2008} considered fixed jackknife limits at $[-\pi/2,+\pi/2]$. We show that the jackknife limits of a typical vehicle-trailer system can be much smaller than $[-\pi/2,+\pi/2]$ and that the jackknife limits vary for different vehicle-trailer systems. Altafini et al. \cite{Altafini2001} and Pradalier and Usher \cite{Pradalier2007} used fixed empirical jackknife limits based upon experiments with the system before operation, which is not desirable. None of these works consider sideslips that could occur in physical systems. The work proposed here can use simple sensors and basic kinematic parameters to characterize slip and estimate varying jackknife limits. 

Our work is based upon analytical kinematic calculations, which some related work have partially considered. González-Cantos and Ollero calculated jackknife limits using vehicle-trailer kinematics, but they only considered on-axle hitching and assumed the jackknife limits to be within $[-\pi/2,+\pi/2]$ \cite{Ollero2009}, but we show that limits can actually be higher. Yuan and Zhu derived another form of jackknife limits for on-axle hitching considering steering limits \cite{Yuan2016}. Chiu and Goswami derived analytical jackknife limits (named ``critical hitch angle") for general vehicle-trailer systems \cite{Goswami2014}. Others have applied similar analytical forms of jackknife limits in their jackknife prevention method \cite{Jing2019} \cite{Aldughaiyem2021}. To the best of our knowledge, the work presented here is the first to examine the impact of sideslip on jackknife limits, and provides a general consideration of vehicle-trailer systems, highlighting two additional jackknife limits not indicated by other authors. 

This research should not be confused with dynamic jackknifing, which is classically the unrecoverable folding of a semitrailer due to braking or a high speed maneuver during driving, which is quite different from ``kinematic jackknifing" considered in this paper. The analysis \cite{VLK1982} and detection  \cite{McCann2005} of dynamic jackknife behavior are done in dynamics. Dynamic jackknife avoidance mainly focuses on maintaining small hitch angle by means of special mechanisms, steering control, or braking control \cite{Keller1973} \cite{Gaussoin1948}. These results are not directly applicable to trailer backing, where kinematic behavior dominates and large hitch angles are frequently used for trailer manipulation.
 
\subsection{Contributions}
This paper derives an analytical form of jackknife limits considering sideslip and various vehicle-trailer configurations, which does not exist in the literature to our knowledge. The work shows that there are four jackknife limits and two non-jackknife regions, which provides detailed consideration of jackknifing. Analysis highlights that there are three trailer lengths that each require special consideration within these regions. The techniques are implemented in simulations in conjunction with trailer backing controllers to demonstrate jackknife limits and behavior with different trailer sizes. Simulations demonstrate improved trailer backing performance when considering the effect of slip on Jackknife limits.  Experiments evaluate the analysis in field operations with a real vehicle, trailer, and sensing system. The methods are shown to apply to both ``front-bumper-hitching" and ``rear-bumper-hitching" configurations, which relate to trailer pushing and trailer backing operations, respectively.  

\subsection{Paper Structure}
Sec.\,\ref{Model} introduces the kinematic model of the vehicle-trailer system. Jackknifing conditions considering sideslip are derived in Sec.\ \ref{jackknife}. Jackknife limit considerations are in Sec.\,\ref{jackknifeImpact}. Simulation and experimental results are in Sec.\,\ref{section_test} with conclusions in Sec.\,\ref{Conclusion}. Nomenclature is listed at the end of the paper.

\section{Vehicle-trailer Kinematic Model}  \label{Model}
The model used in this paper is derived in \cite{Leng2016}, which is is presented concisely since it is used extensively in this paper. The geometry of a vehicle-trailer system is shown in Fig. \ref {fig:model}, where symbols are indicated in the Nomenclature. Angles are shown in their positive regions. Note that front wheel sideslip angle, $\beta_F$, is defined as the speed direction of the front wheel minus the facing direction of the front wheel. Vehicle curvature, $\kappa$, is the instantaneous movement curvature of the vehicle, which is a function of vehicle steering angle as shown in (\ref{kappaV}). The upper vehicle curvature limit, $\kappa^{}_{\max}$, may equal $+\infty$, which means the analysis considers a vehicle that can turn left with zero turning radius like a unicycle type robot, fork-lift, or skid-steering vehicle. Similarly, the lower vehicle curvature limit, $\kappa^{}_{\min}$, may equal $-\infty$. The extended kinematic model of the vehicle-trailer system including sideslip is then,
\begin{eqnarray}
\dot x &=& v\cos (\theta_V + \beta_R)
\label{modelr1}\\
\dot y &=& v\sin (\theta_V + \beta_R)
\label{modelr2}\\
\dot \theta_V  &=& v\kappa
\label{modelr3}\\
\dot \psi  &=&  - v\left( \kappa + \frac{{\sin (\psi -\beta_R +\beta_T)}}{{L_2 \cos\beta_T}}  + \frac{{L_1 \kappa \cos (\psi +\beta_T)}}{{L_2 \cos\beta_T}} \right)
\label{modelr4}
\end{eqnarray}
\noindent where the vehicle curvature is
\begin{eqnarray}
\kappa=(\tan(\phi+\beta_F)\cos\beta_R-\sin\beta_R)/L.
\label{kappaV}
\end{eqnarray}

In the case of the unicycle type robot where curvature ranges are $[-\infty,+\infty]$, it is helpful to replace $v\kappa$ coefficients in (\ref{modelr3}) and (\ref{modelr4}) equivalently with $\omega$, the heading rate of the robot.  That resolves numerical issues associated with $\kappa = \pm \infty$ where velocity is zero, while still allowing the following mathematical analysis to be applied.

Sideslip angles are considered as exogenous variables, which can either be estimated or measured. Since this paper is focused on the analysis of hitch angle change based on kinematics and zero vehicle velocity results in zero hitch angle change in kinematics analysis, vehicle velocity, $v$, is generally considered non-zero throughout this paper.

\section {Jackknife Analysis} \label{jackknife}

Jackknife criteria and the analytical form of jackknife limits are now derived. Jackknife regions and non-jackknife regions are analyzed. It is assumed that velocity sign is always maintained. It is also assumed that sideslip variations are slow, which is reasonable for typical slow trailer operations.

\subsection{Jackknife Criteria} \label{Sec_JK_Critearia}

According to earlier discussion, the jackknife state can be defined as follows.

\begin{MyDef}
A jackknife state is a vehicle-trailer system state where the sign of hitch angle rate remains constant regardless of the achievable vehicle curvature applied. 
\label{DefJK}
\end{MyDef}

In other words, if a vehicle-trailer system is in a jackknife state, no matter what steering angle the user selects, they cannot stop hitch angle from changing in one direction, e.g., magnitude continues to increase or decrease. Theorem \ref{TheoJKCriteria} can be used to determine whether a system state is a jackknife state.

\begin{MyTheo}
Given sideslip, vehicle velocity, and hitch angle,
\begin{enumerate}
\item If $L_2\cos\beta_T+L_1\cos(\psi+\beta_T)=0$, then the system is in a jackknife state.
\item If $\dot\psi=0$ cannot be achieved by achievable vehicle curvatures and $L_2\cos\beta_T+L_1\cos(\psi+\beta_T)\ne 0$, then the system is in a jackknife state.
\item For all other cases $\dot\psi=0$ can be achieved and the system is in a non-jackknife state.
\end{enumerate}

\label{TheoJKCriteria}
\end{MyTheo}

\begin{MyProof}
For the first case, according to (\ref{modelr4}), if $L_2\cos\beta_T+L_1\cos(\psi+\beta_T)=0$, terms related to vehicle curvature $\kappa$, or equivalently $v\kappa=\omega$, would cancel each other. Then all vehicle curvatures or angular rates would result in the same $\dot\psi$. According to Definition \ref{DefJK}, the system is in a jackknife state.

For the second case, according to (\ref{modelr4}), when hitch angle, sideslip, and velocity are given and  $L_2\cos\beta_T+L_1\cos(\psi+\beta_T)\ne 0$, hitch angle rate $\dot\psi$ is continuous and strictly monotonic with respect to vehicle curvature $\kappa$. If $\dot\psi=0$ cannot be achieved by achievable vehicle curvatures (i.e. no zero crossing), then all $\dot\psi$ achieved by achievable vehicle curvatures would be of the same sign, i.e. either positive or negative. According to Definition \ref{DefJK}, the system is in a jackknife state.

For the third case, $\dot\psi=0$ can be achieved by an achievable vehicle curvature and $L_2\cos\beta_T+L_1\cos(\psi+\beta_T)\ne 0$. According to (\ref{modelr4}), when hitch angle, sideslip, and velocity are given and  $L_2\cos\beta_T+L_1\cos(\psi+\beta_T)\ne 0$, hitch angle rate, $\dot\psi$, is continuous and strictly monotonic with respect to vehicle curvature, $\kappa$. If $\dot\psi=0$ can be achieved by an achievable vehicle curvature, then either $\dot\psi\ge 0$ or $\dot\psi\le 0$ (or both of them) can be achieved by achievable vehicle curvatures, too. According to Definition \ref{DefJK}, the system is in a non-jackknife state.
\qedsymbol
\end{MyProof}

The two criteria in Theorem \ref{TheoJKCriteria} that result in jackknifing are analyzed to gain insight about how steering limits affect jackknifing. Based upon (\ref{modelr4}), the two hitch angles satisfying $L_2\cos\beta_T+L_1\cos(\psi+\beta_T)=0$ are,
\begin{eqnarray}
\psi^{+}_{\infty} &=& +\arccos(-L_2\cos\beta_T/L_1) - \beta_T  \label{psiR_Infty}\\
\psi^{-}_{\infty} &=& -\arccos(-L_2\cos\beta_T/L_1) - \beta_T. \label{psiL_Infty}
\end{eqnarray}

\noindent where  $\psi^{+}_{\infty}$ and $\psi^{-}_{\infty}$, defined as \textbf{Uncontrollable Hitch Angles}, exist when $|L_2\cos\beta_T/L_1|\le 1$, i.e. $L_2\le|L_1/\cos\beta_T|$. Note that $L_2>0$ by definition.  To further analyze the second and third criterion related to $\dot\psi=0$, we define the \textbf{Critical Vehicle Curvature}, $\kappa^*_{}(\psi)$,  based upon (\ref{modelr4}) that makes $\dot\psi=0$,

\begin{eqnarray}
\kappa^*_{}(\psi)={\frac{-\sin(\psi-\beta_R+\beta_T)}{L_2\cos\beta_T+L_1\cos(\psi+\beta_T)}}.
\label{K_V0}
\end{eqnarray}

\noindent By definition, $\kappa^*_{}(\psi)$ exists if and only if $\psi\ne\psi^{+}_{\infty}$ and $\psi\ne\psi^{-}_{\infty}$. If $\kappa^*_{}$ exists and $\kappa^*_{}(\psi)\notin[\kappa^{}_{\min},\kappa^{}_{\max}]$, then $\dot\psi=0$ cannot be achieved by achievable vehicle curvatures.  Given $\kappa^*_{}(\psi)$, we can rewrite Theorem \ref{TheoJKCriteria}:

\begin{figure}
\centerline{\epsfig{figure=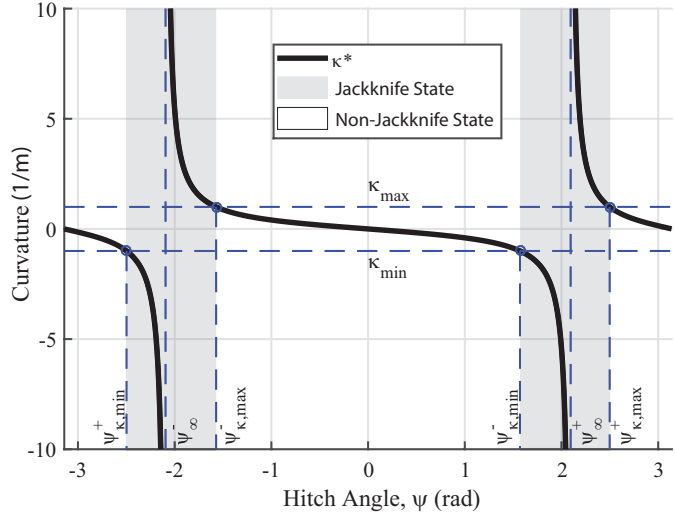,height=6.85cm}}
\vspace{-.1cm}
\caption{Required vehicle curvatures for maintaining hitch angles for a Short Trailer with $L_1>0$.}
\label{fig:kappaV0_Short}
\end{figure}

\begin{MyLemma} \label{LemmaJKCriteria}
With given sideslip, vehicle velocity, and hitch angle, $\psi_1$, 
\begin{enumerate}
\item If $\psi_1=\psi^{-}_{\infty}$, $\psi_1=\psi^{+}_{\infty}$, or $\kappa^*_{}(\psi_1)\notin[\kappa^{}_{\min},\kappa^{}_{\max}]$, then the system is in a \textbf{jackknife state}.
\item Otherwise, the system is in a \textbf{non-jackknife state}.
\end{enumerate}
\end{MyLemma}

\begin{MyProof}
If $\psi=\psi^{-}_{\infty}$ or $\psi=\psi^{+}_{\infty}$, then $L_2\cos\beta_T+L_1\cos(\psi+\beta_T)=0$. According to Theorem \ref{TheoJKCriteria}, this is a jackknife state.

If $\psi_1\ne\psi^{-}_{\infty}$, $\psi_1\ne\psi^{+}_{\infty}$, and $\kappa^*_{}(\psi_1)\notin[\kappa^{}_{\min},\kappa^{}_{\max}]$, then $\dot\psi=0$ cannot be achieved by achievable vehicle curvatures and $L_2\cos\beta_T+L_1\cos(\psi+\beta_T)\ne 0$, which, according to Theorem \ref{TheoJKCriteria}, means the system is in a jackknife state.
\qedsymbol
\end{MyProof}

 An illustration of Critical Vehicle Curvature $\kappa^*_{}(\psi)$ required to achieve $\dot\psi=0$ is shown in Fig.\,\ref{fig:kappaV0_Short}. $\kappa^{}_{\min}$ and $\kappa^{}_{\max}$ indicate vehicle curvature limits leading to jackknife and non-jackknife states. Critical vehicle curvature is discontinuous at the uncontrollable hitch angles $\psi^{+}_{\infty}$ and $\psi^{-}_{\infty}$. 


\subsection{Jackknife Limits} \label{JK_Limits}

Critical Hitch Angle is now introduced as a tool for determining jackknife limits.  

\begin{MyDef}
A hitch angle, $\psi_1$, is a \textbf{Critical Hitch Angle} if and only if its Critical Vehicle Curvature satisfies $\kappa^*_{}(\psi_1)=\kappa^{}_{\min}$ or $\kappa^*_{}(\psi_1)=\kappa^{}_{\max}$.
\label{Def_Critical_HitchAngle}
\end{MyDef}

The upper and lower limits of the jackknife state regions can be defined as jackknife limits. Due to the special property of the Critical Hitch Angles (i.e., at these hitch angles, the system is able to maintain hitch angle and move hitch angle in only one direction), the Critical Hitch Angles within a non-jackknife region are still defined as jackknife limits for the ease of later analysis. Thus jackknife limits are defined as follows.
\begin{MyDef}
The upper and lower limits of the jackknife state regions and the Critical Hitch Angles within non-jackknife regions are defined as the \textbf{Jackknife Limits}.
\label{DefJK2}
\end{MyDef}

Using the theorem and lemma above, we can test a given hitch angle for jackknife limit. We can easily test a given hitch angle for jackknife limit if it does not equal $\psi^{+}_{\infty}$ and $\psi^{-}_{\infty}$ using the following lemma.

\begin{MyLemma} \label{LemmaJKLimit}
For a given sideslip and velocity, a hitch angle, $\psi_1$, satisfying $\psi_1\ne\psi^{+}_{\infty}$ and $\psi_1\ne\psi^{-}_{\infty}$ is a jackknife limit if and only if it is a Critical Hitch Angle.
\end{MyLemma}

\begin{MyProof}
Let $\psi_1$ be a hitch angle not equal to $\psi^{+}_{\infty}$ nor $\psi^{-}_{\infty}$.

According to (\ref{modelr4}), if sideslip and vehicle velocity are given, then hitch angle rate at $\psi_1$, $\dot\psi(\psi_1)$, is continuous and strictly monotonic with respect to vehicle curvature, $\kappa$. The derivative of $\kappa^*_{}(\psi)$ with respect to $\psi$ is
\begin{eqnarray}
{\frac{\partial \kappa^*_{}}{\partial \psi}} =  - {\frac{L_2\cos(\psi-\beta_R+\beta_T)\cos\beta_T+L_1\cos\beta_R}{(L_2\cos\beta_T + L_1\cos(\psi+\beta_T))^2}}.
\label{dKappa_dPsi}
\end{eqnarray}

\noindent{According to (\ref{dKappa_dPsi}), ${\partial \kappa^*_{}}/{\partial \psi}$ can be positive, negative, or zero. Let $\psi_{1L}$ be the hitch angle point adjacent to and on the left side of $\psi_1$ (i.e. less than $\psi_1$). Let $\psi_{1R}$ be the hitch angle point adjacent to and on the right side of $\psi_1$ (i.e. greater than $\psi_1$).}

Assume ${\partial \kappa^*_{}}/{\partial \psi}>0$ at $\psi_1$. If $\kappa^*_{}(\psi_1)=\kappa^{}_{\min}$, then $\kappa^*_{}(\psi_{1L})<\kappa^{}_{\min}$ and $\kappa^*_{}(\psi_{1R})>\kappa^{}_{\min}$. According to Lemma \ref{LemmaJKCriteria}, $\psi_{1L}$ is jackknifing while $\psi_{1R}$ is not. By Definition \ref{DefJK2}, $\psi_1$ is a jackknife limit. Similarly, it can be proven that if $\kappa^*_{}(\psi_1)=\kappa^{}_{\max}$, then $\psi_1$ is a jackknife limit. Thus, Critical Hitch Angles are jackknife limits when ${\partial \kappa^*_{}}/{\partial \psi}>0$.

If $\kappa^*_{}(\psi_1)\in(\kappa^{}_{\min},\kappa^{}_{\max})$, then $\psi_{1L}$ and $\psi_{1R}$ are all non-jackknifing and thus $\psi_{1}$ is not a jackknife limit. If $\kappa^*_{}(\psi_1)\notin [\kappa^{}_{\min},\kappa^{}_{\max}]$, then $\psi_1$ is jackknifing and is not a jackknife limit. Thus Lemma \ref{LemmaJKLimit} is proven on the condition of ${\partial \kappa^*_{}}/{\partial \psi}>0$ at $\psi_1$, which is similar for ${\partial \kappa^*_{}}/{\partial \psi}<0$ at $\psi_1$.

In the rare case of ${\partial \kappa^*_{}}/{\partial \psi}=0$ at $\psi_1$, then $\psi_{1L}$ and $\psi_{1R}$ are either all non-jackknifing or all jacknifing. If both $\psi_{1L}$ and $\psi_{1R}$ are jacknifing, since $\psi_1$ is not jackknifing, then $\psi_1$ is apparently a boundary between jackknifing and non-jackknifing regions. According to Definition \ref{DefJK2}, $\psi_1$ is a jackknife limit. If both $\psi_{1L}$ and $\psi_{1R}$ are non-jacknifing, $\psi_1$ is a Critical Hitch Angle within a non-jackknifing region. According to Definition \ref{DefJK2}, $\psi_1$ is also a jackknife limit.
\qedsymbol
\end{MyProof}

According to Lemma \ref{LemmaJKLimit}, it can be calculated that there are four jackknife limits, which are denoted as $\psi^{+}_{\kappa,\max}$, $\psi^{-}_{\kappa,\max}$, $\psi^{+}_{\kappa,\min}$ and $\psi^{-}_{\kappa,\min}$. Among them, $\psi^{+}_{\kappa,\max}$ and $\psi^{-}_{\kappa,\max}$ are the jackknife limits satisfying $\kappa^*_{}(\psi^{+}_{\kappa,\max})=\kappa^*_{}(\psi^{-}_{\kappa,\max})=\kappa^{}_{\max}$. Similarly, $\psi^{+}_{\kappa,\min}$ and $\psi^{-}_{\kappa,\min}$ are the jackknife limits satisfying $\kappa^*_{}(\psi^{+}_{\kappa,\min})=\kappa^*_{}(\psi^{-}_{\kappa,\min})=\kappa^{}_{\min}$. Four \textit {slip-based} \textbf{Critical Hitch Angles} can then be derived from (\ref{K_V0}) ,
\begin{eqnarray}
\psi^{+}_{\kappa,\max}&=&+\alpha_{1}(\kappa^{}_{\max})+\alpha_{2}(\kappa^{}_{\max}) -\beta_T \label{psiMaxIn}\\
\psi^{-}_{\kappa,\max}&=&-\alpha_{1}(\kappa^{}_{\max})+\alpha_{2}(\kappa^{}_{\max}) -\beta_T \label{psiMaxOut}\\
\psi^{+}_{\kappa,\min}&=&+\alpha_{1}(\kappa^{}_{\min}) + \alpha_{2}(\kappa^{}_{\min})  -\beta_T  \label{psiMinIn}\\
\psi^{-}_{\kappa,\min}&=&-\alpha_{1}(\kappa^{}_{\min}) + \alpha_{2}(\kappa^{}_{\min})  -\beta_T \label{psiMinOut}
\end{eqnarray}

\noindent where $\alpha_1$ and $\alpha_2$ are,
\begin{eqnarray}
\alpha_{1}(\kappa)&=&\arccos\left(\frac{-L_2\cos\beta_T\kappa^{}_{}}{\sqrt{L^2_1 \kappa^2_{}-2\sin\beta_R L_1 \kappa^{}_{} +1}}\right) \label{alpha1}\\
\alpha_{2}(\kappa)&=&\arctan 2(\cos\beta_R, L_1 \kappa^{}_{}-\sin\beta_R). \label{alpha2}
\end{eqnarray}

\noindent Critical Hitch Angles are typically Jackknife Limits, although they may not all exist since the $\arccos()$ in $\alpha_1$ may not have a real solution, which is discussed in later subsections.

Then, $\psi^{+}_{\infty}$ and $\psi^{-}_{\infty}$ are tested for jackknife limits. They can be easily tested using the following lemma.

\begin{MyLemma} \label{LemmaJKLimit2}
$\psi^{+}_{\infty}$ is a jackknife limits when and only when it equals one of the four critical hitch angles (\ref{psiMaxIn}) to (\ref{psiMinOut}). So is $\psi^{-}_{\infty}$.
\end{MyLemma}

\begin{MyProof}
As mentioned earlier, at $\psi^{+}_{\infty}$ and $\psi^{-}_{\infty}$, vehicle curvature in (\ref{modelr4}) is cancelled and hitch angle rate is constant and uncorrelated to vehicle curvature. Thus, per Definition \ref{DefJK}, these points are always jackknifing. Typically, these points, as long as they exist, are within jackknife regions and not adjacent to a non-jackknife region and thus are not jackknife limits, Fig.\,\ref{fig:kappaV0_Short}. 

The only condition that $\psi^{+}_{\infty}$ (or $\psi^{-}_{\infty}$) is a jackknife limit is that one of the hitch angle value adjacent to it is a non-jackknife hitch angle. Since $\psi^{+}_{\infty}$ and $\psi^{-}_{\infty}$ are not adjacent to each other, if $\psi_1$ is a hitch angle adjacent to either of them, then $\psi_1\ne\psi^{+}_{\infty}$ and $\psi_1\ne\psi^{-}_{\infty}$.

According to Lemma \ref{LemmaJKLimit}, for $\psi_1$ to be a non-jackknife hitch angle, there must be $\kappa^*_{}(\psi_1)\in[\kappa^{}_{\min},\kappa^{}_{\max}]$, i.e. $\kappa^*_{\psi_1}$ exists in such case.

Note that the left-hand and right-hand limits of $\kappa^*_{}(\psi^{+}_{\infty})$ are $-\infty$ and $+\infty$, or $+\infty$ and $-\infty$, respectively. It is similar for the left-hand and right-hand limits of $\kappa^*_{}(\psi^{-}_{\infty})$. This means one of the hitch angles adjacent to $\psi^{+}_{\infty}$ and $\psi^{-}_{\infty}$ are non-jackknifing if one of the achievable vehicle curvature limit is unlimited (i.e., $\kappa^{}_{\min} = -\infty$ or $\kappa^{}_{\max} = +\infty$). Thus, $\psi^{-}_{\infty}$ and $\psi^{+}_{\infty}$ are jackknife limits if and only if $\kappa^{}_{\min} = -\infty$ or $\kappa^{}_{\max} = +\infty$.

Furthermore, it can be found that when substituting $\kappa^{}_{\min} = -\infty$ and/or $\kappa^{}_{\max} = +\infty$ into (\ref{psiMaxIn}--\ref{psiMinOut}), $\psi^{-}_{\infty}$ and $\psi^{+}_{\infty}$ can be correctly calculated as jackknife limits. Thus, the jackknife limits at $\psi^{+}_{\infty}$ and $\psi^{-}_{\infty}$ do not need to be considered specifically when calculating jackknife limits.
\qedsymbol
\end{MyProof}

Combining Lemma \ref{LemmaJKLimit} and Lemma \ref{LemmaJKLimit2}, the locations of jackknife limits can be summarized as follows.

\begin{MyLemma} \label{LemmaJKLimit3}
The jackknife limits of a vehicle-trailer system are $\psi^{+}_{\kappa,\max}$, $\psi^{-}_{\kappa,\max}$, $\psi^{+}_{\kappa,\min}$, and $\psi^{-}_{\kappa,\min}$.
\end{MyLemma}

Note that there is a corresponding achievable vehicle curvature limit for each possible jackknife limit, Fig.\,\ref{fig:kappaV0_Short}, while there are two possible jackknife limits corresponding to each achievable vehicle curvature limit. The corresponding achievable vehicle curvature limit for $\psi^{+}_{\kappa,\max}$ and $\psi^{-}_{\kappa,\max}$ is $\kappa^{}_{\max}$. The corresponding achievable vehicle curvature limit for $\psi^{+}_{\kappa,\min}$ and $\psi^{-}_{\kappa,\min}$ is $\kappa^{}_{\min}$. 


\subsection{Vehicle-Trailer Categories} \label{TrailerCategories}

According to Lemma \ref{LemmaJKLimit3}, there are up to four jackknife limits, which separate the range of $\psi$ into up to two jackknife regions and two non-jackknife regions. To further analyze these limits and regions, vehicle-trailer systems are divided into three categories: Short, Medium, and Long Trailer, Table \ref{tab:V_T_Categories}.
\begin{table}
\caption{Vehicle-Trailer Categories} \label{tab:V_T_Categories}
\begin{center}
\begin{tabular}{| l |  l | }
\hline
Type & Criteria\\
\hline                   
Short Trailer& $L_2\in(0,|L_1\cos\beta_R\sec\beta_T|]$ \\
\hline              
Medium Trailer&$L_2\in(|L_1\cos\beta_R\sec\beta_T|,|L_1\sec\beta_T|]$\\
\hline
Long Trailer&$L_2\in(|L_1\sec\beta_T|,+\infty)$\\
\hline
\end{tabular}
\end{center}
\end{table}

The criterion dividing Short and Medium Trailer, i.e. $L_2\le |L_1\cos\beta_R\sec\beta_T|$, is based on (\ref{dKappa_dPsi}), which is a criterion for the existence of local minimum and maximum of Critical Vehicle Curvature. The criterion dividing Medium and Long Trailer, i.e. $L_2>|L_1\sec\beta_T|$, is based on (\ref{K_V0}), which is a criterion for bounded Critical Vehicle Curvature. On-axle hitching (i.e., $L_1=0$) causes all trailers to be a ``Long Trailer," since trailer tongue length, $L_2$, is always larger than zero. These three categories of vehicle-trailer systems are analyzed in detail below.


\subsection{Short Trailer Analysis} \label{shortTrailer}
Per (\ref{K_V0}), the general relation between Critical Vehicle Curvature, $\kappa^*_{}$, and hitch angle, $\psi$, for a Short Trailer is shown in Fig.\,\ref{fig:kappaV0_Short} ($L_1=2$\,m, $L_2=1$\,m, $\beta_T=0^\circ$, $\beta_R=0^\circ$, $\kappa^{}_{\max}=1.761$\,m$^{-1}$, $\kappa^{}_{\min}=-1.761$\,m$^{-1}$) for $L_1>0$. When $L_1<0$, the hitch point is in front of the rear axle, which can occur with trailer pushing when the hitch is on the front bumper; the general shape of the curve is mirrored with respect to the $\psi$ axis. $L_1=0$ is not considered since this subcase results in $L_2=0$ according to the Short Trailer criteria in Table \ref{tab:V_T_Categories}, which is unrealistic.

For checking the existence of the jackknife limits, the $\arccos()$ operand in (\ref{alpha1}) is examined. Considering the Short Trailer criteria, it can be found that
\begin{eqnarray}
& &\sqrt{L^2_1 \kappa^2_{}-2\sin\beta_R L_1 \kappa^{}_{} +1} \nonumber\\
&=&\sqrt{(L_1\kappa_{}\sin\beta_R-1)^2+L^2_1\kappa^2_{}\cos^2\beta_R} \nonumber\\
&\ge&\sqrt{L^2_1\kappa^2_{}\cos^2\beta_R} \nonumber\\
&\ge&\sqrt{L^2_2\kappa^2_{}\cos^2\beta_T}=|L_2\kappa\cos\beta_T|.
\end{eqnarray}

\noindent As a result of this inequality, it is always true that
\begin{eqnarray}
\left|\frac{L_2\kappa\cos\beta_T}{\sqrt{L^2_1 \kappa^2_{}-2\sin\beta_R L_1 \kappa^{}_{} +1}} \right|\le 1.
\end{eqnarray}

\noindent Therefore, according to (\ref{alpha1}), the $\arccos()$ operand in $\alpha_1$ is always within $[-1,+1]$, which means that all jackknife limits exist for the Short Trailers.

In this case there are two isolated non-jackknife regions: one is between $\psi^{-}_{\kappa,\max}$ and $\psi^{-}_{\kappa,\min}$; the other is between and $\psi^{+}_{\kappa,\max}$ and $\psi^{+}_{\kappa,\min}$, usually across $\pm\pi$\,rad. The first non-jackknife region, which is usually around zero hitch angle, is desired in most applications. The second non-jackknife region, which is usually around $\pm\pi$\,rad hitch angle, is desired only in some special applications, for example pushing a trailer with a front bumper hitching configuration \cite{d3}. In rear hitch systems, hitch angle cannot reach this non-jackknife region due to physical (e.g. mechanical) limitations.

The non-jackknife regions and jackknife limits (i.e., the boundaries of the non-jackknife regions) of the Short Trailer case can be derived per the following subcases:
\begin{enumerate}
\item{Subcase S-1:} if $L_1>0$, there are two non-jackknife regions. The first non-jackknife region is $[\psi^{-}_{\kappa,\max},\psi^{-}_{\kappa,\min}]$. The second non-jackknife region is $[\psi^{+}_{\kappa,\max},\psi^{+}_{\kappa,\min}]$.

\item{Subcase S-2:} if $L_1<0$, there are two non-jackknife regions. The first non-jackknife region is $[\psi^{-}_{\kappa,\min},\psi^{-}_{\kappa,\max}]$. The second non-jackknife region is $[\psi^{+}_{\kappa,\min},\psi^{+}_{\kappa,\max}]$.
\end{enumerate}

\subsection{Medium Trailer Analysis} \label{mediumTrailer}
\begin{figure}
\centerline{\epsfig{figure=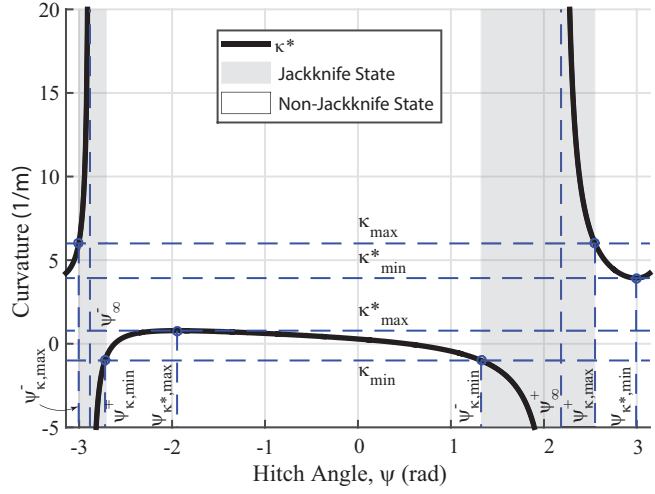,height=6.55cm}}
\vspace{-0.1 cm}
\caption{Required vehicle curvatures for maintaining hitch angles for a Medium Trailer with $L_1>0$ for subcase M-3.}
\label{fig:kappaV0_Medium}
\end{figure}

Per (\ref{K_V0}), the general relation between Critical Vehicle Curvature, $\kappa^*_{}$, and hitch angle, $\psi$, for a Medium Trailer is shown in Fig.\,\ref{fig:kappaV0_Medium} ($L_1=1m$, $L_2=0.8741m$, $\beta_T=20^\circ$, $\beta_R=50^\circ$, $\kappa^{}_{\max}=6m^{-1}$, $\kappa^{}_{\min}=-1m^{-1}$) for $L_1>0$. Parameters are selected so as to make the relative position of all the $\kappa$ and $\psi$ easy to read. When $L_1<0$, the general shape of the curve is mirrored with respect to the $\psi$ axis. $L_1=0$ is not considered since this results in $L_2=0$ per Table \ref{tab:V_T_Categories}, which is unrealistic. 

This plot is not symmetric because sideslip angles, i.e., $\beta_R$ and $\beta_T$, are not zero, which skews the plot. The Medium Trailer case is caused by nonzero rear wheel sideslip, $\beta_R$. If $\beta_R=0$, then there would only be Short Trailer and Long Trailer cases.

According to (\ref{dKappa_dPsi}) and the Medium Trailer criteria, the vehicle curvature required to maintain a given hitch angle, $\kappa^*_{}(\psi)$, is not monotonic with respect to hitch angle, $\psi$, for the Medium Trailer. There is a local maximum and a local minimum of $\kappa^*_{}(\psi)$ at $\psi_{\kappa *,\max}$ and $\psi_{\kappa *,\min}$ respectively, which can be derived from (\ref{dKappa_dPsi}) as
\begin{eqnarray}
\psi_{\kappa *,\max}=-\arccos\left(\frac{-L_1\cos\beta_R}{L_2\cos\beta_T}\right)+\beta_R-\beta_T \label{psiKappaMax}\\
\psi_{\kappa *,\min}=+\arccos\left(\frac{-L_1\cos\beta_R}{L_2\cos\beta_T}\right)+\beta_R-\beta_T. \label{psiKappaMin}
\end{eqnarray}

$\kappa^{*}_{\max}$ and $\kappa^{*}_{\min}$ are accordingly the local maximum and local minimum of  $\kappa^*_{}(\psi)$, which can be derived by substituting (\ref{psiKappaMax}) and (\ref{psiKappaMin}) into (\ref{K_V0}) separately as
\begin{eqnarray}
\kappa^{*}_{\max} = \frac{L_1\sin\beta_R-\sqrt{L^2_2\cos^2\beta_T-L^2_1\cos^2\beta_R}}{L^2_1-L^2_2\cos^2\beta_T} \label{kappaV0Max}\\
\kappa^{*}_{\min} = \frac{L_1\sin\beta_R+\sqrt{L^2_2\cos^2\beta_T-L^2_1\cos^2\beta_R}}{L^2_1-L^2_2\cos^2\beta_T}.\label{kappaV0Min}
\end{eqnarray}
\noindent Note that $\kappa^{*}_{\max}<\kappa^{*}_{\min}$ for the Medium Trailer case.

At the upper boundary of the Medium Trailer criteria, i.e., $L_2=|L_1\sec\beta_T|$, $\kappa^{*}_{\max}$ and $\kappa^{*}_{\min}$ can be derived using the left-hand limit $L_2 \to |L_1\sec\beta_T|^-$ and L'Hopital's rule. Thus, for $L_1>0$ and $L_2=|L_1\sec\beta_T|$, it can be derived via (\ref{kappaV0Max}) and (\ref{kappaV0Min}) that $\kappa^{*}_{\max}=1/(2L_1\sin\beta_R)$ and $\kappa^{*}_{\min}=+\infty$. Similarly, for $L_1<0$ and $L_2=|L_1\sec\beta_T|$, it can be derived that $\kappa^{*}_{\max}=-\infty$ and $\kappa^{*}_{\min}=1/(2L_1\sin\beta_R)$.

The jackknife limits of Medium Trailers may not always exist. It can be shown from (\ref{psiMaxIn}), (\ref{psiMaxOut}), and (\ref{alpha1}) that $\psi^{-}_{\kappa,\max}$ and $\psi^{+}_{\kappa,\max}$ exist when $\alpha_1(\kappa^{}_{\max})$ is real. Similarly $\psi^{+}_{\kappa,\min}$ and $\psi^{-}_{\kappa,\min}$ exist when $\alpha_1(\kappa^{}_{\min})$ is real. According to (\ref{alpha1}), for $\alpha_1(\kappa)$ to be real, $\kappa$ must satisfy $L^2_1 \kappa^2_{}-2\sin\beta_R L_1 \kappa^{}_{} +1\ge L^2_2\cos^2 \beta_T\kappa^2_V$, which can be written as
\begin{eqnarray}
(L^2_1 - L^2_2\cos^2 \beta_T)\kappa^2_{}-2\sin\beta_R L_1 \kappa^{}_{} +1\ge 0. \label{existanceCondition}
\end{eqnarray}

The two boundary points solved from (\ref{existanceCondition}) are exactly $\kappa^{*}_{\max}$ and $\kappa^{*}_{\min}$ as in (\ref{kappaV0Max}) and (\ref{kappaV0Min}). According to the Medium Trailer criteria shown in Table \ref{tab:V_T_Categories}, $L^2_1 - L^2_2\cos^2 \beta_T\ge0$. Therefore, the $\kappa$ satisfying (\ref{existanceCondition}) (i.e., resulting in real $\alpha_1$) can be solved as $\kappa \le \kappa^{*}_{\max}$ or $\kappa \ge \kappa^{*}_{\min}$. 

Thus, $\psi^{-}_{\kappa,\max}$ and $\psi^{+}_{\kappa,\max}$ exist when $\kappa^{}_{\max} \notin (\kappa^{*}_{\max},\kappa^{*}_{\min})$, and $\psi^{+}_{\kappa,\min}$ and $\psi^{-}_{\kappa,\min}$ exist when $\kappa^{}_{\min} \notin (\kappa^{*}_{\max},\kappa^{*}_{\min})$. This conclusion can also be drawn intuitively from Fig.\,\ref{fig:kappaV0_Medium}.

Considering $\kappa^{*}_{\max}< \kappa^{*}_{\min}$, the non-jackknife regions and jackknife limits of the Medium Trailer case can be derived per the following subcases for $L_1>0$ (i.e., rear-bumper-hitching), 
\begin{enumerate}
\item{Subcase M-1:} if $\kappa^{}_{\max}\ge\kappa^{*}_{\min}$ and $\kappa^{}_{\min}\ge\kappa^{*}_{\min}$, there are two non-jackknife regions. The first non-jackknife region is $[\psi^{-}_{\kappa,\min},\psi^{-}_{\kappa,\max}]$. The second non-jackknife region is $[\psi^{+}_{\kappa,\max}, \psi^{+}_{\kappa,\min}]$.


\item{Subcase M-2:} if $\kappa^{}_{\max}\ge\kappa^{*}_{\min}$ and $\kappa^{}_{\min}\in (\kappa^{*}_{\max},\kappa^{*}_{\min})$, there is only one non-jackknife region, which is $[\psi^{+}_{\kappa,\max},\psi^{-}_{\kappa,\max}]$.


\item{Subcase M-3:} if $\kappa^{}_{\max}\ge\kappa^{*}_{\min}$ and $\kappa^{}_{\min}\le\kappa^{*}_{\max}$, there are two non-jackknife regions. The first non-jackknife region is $[\psi^{+}_{\kappa,\max},\psi^{-}_{\kappa,\max}]$. The second non-jackknife region is $[\psi^{+}_{\kappa,\min}, \psi^{-}_{\kappa,\min}]$. This is shown in Fig.\,\ref{fig:kappaV0_Medium}.


\item{Subcase M-4:} if $\kappa^{}_{\max}\in (\kappa^{*}_{\max},\kappa^{*}_{\min})$ and $\kappa^{}_{\min}\in (\kappa^{*}_{\max},\kappa^{*}_{\min})$, then their is no non-jackknife region and no jackknife limit since jackknifing occurs under all conditions.

\item{Subcase M-5:} if $\kappa^{}_{\max}\in (\kappa^{*}_{\max},\kappa^{*}_{\min})$ and $\kappa^{}_{\min}\le\kappa^{*}_{\max}$, there is only one non-jackknife region, which is $[\psi^{+}_{\kappa,\min},\psi^{-}_{\kappa,\min}]$.


\item{Subcase M-6:} if $\kappa^{}_{\max}\le\kappa^{*}_{\max}$ and $\kappa^{}_{\min}\le\kappa^{*}_{\max}$, there are two non-jackknife regions. The first non-jackknife region is $[\psi^{-}_{\kappa,\max},\psi^{-}_{\kappa,\min}]$. The second non-jackknife region is $[\psi^{+}_{\kappa,\min},\psi^{+}_{\kappa,\max}]$.

\end{enumerate}

Similar subcases occur for $L_1<0$, which are left to the reader, where non-jackknife region upper and lower boundaries are switched.

\subsection{Long Trailer Analysis} \label{longTrailer}
\begin{figure}
\centerline{\epsfig{figure=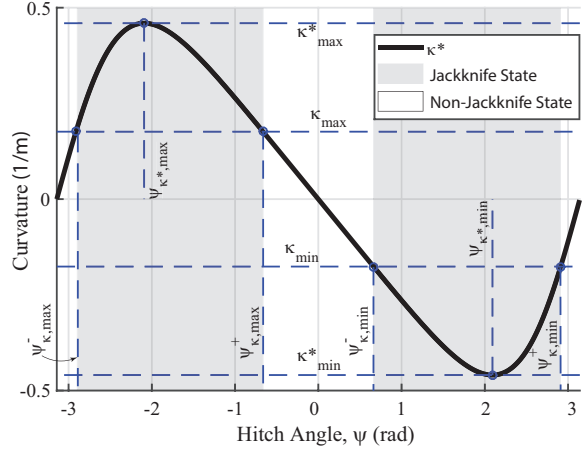,height=6cm}}
\vspace{-0.1 cm}
\caption{Required vehicle curvatures for maintaining hitch angles for a Long Trailer with $L_1>0$. Subcase L-4 is shown.}
\label{fig:kappaV0_Long}
\end{figure}

The Long Trailer case is the most common case in trailer applications. Per (\ref{K_V0}), the relation between Critical Vehicle Curvature, $\kappa^*_{}$, and hitch angle, $\psi$, of a Long Trailer is shown in Fig.\,\ref{fig:kappaV0_Long} ($L_1=1.23m$, $L_2=2.51m$, $\beta_T=0^\circ$, $\beta_R=0^\circ$, $\kappa^{}_{\max}=0.1761m^{-1}$, $\kappa^{}_{\min}=-0.1761m^{-1}$).

For Long Trailers, $\psi_{\kappa *,\max}$, $\psi_{\kappa *,\min}$, $\kappa^{*}_{\max}$, and $\kappa^{*}_{\min}$ have the same definition and expressions as (\ref{psiKappaMax}--\ref{kappaV0Min}). Note that the local maximum of $\kappa^*_{}$, $\kappa^{*}_{\max}$, is greater than the local minimum, $\kappa^{*}_{\min}$, in this case due to the Long Trailer criteria.

The proof of existence of the jackknife limits is similar to that of the Medium Trailer case. However, the difference is that due to the Long Trailer criteria shown in Table \ref{tab:V_T_Categories}, $L^2_1 - L^2_2\cos^2 \beta_T<0$. Similar to the prior section, it can be shown that $\psi^{-}_{\kappa,\max}$ and $\psi^{+}_{\kappa,\max}$ exist only when $\kappa^{}_{\max}\in[\kappa^{*}_{\min},\kappa^{*}_{\max}]$, which is left for reader to confirm. Likewise, $\psi^{+}_{\kappa,\min}$ and $\psi^{-}_{\kappa,\min}$ exist only when $\kappa^{}_{\min}\in[\kappa^{*}_{\min},\kappa^{*}_{\max}]$. 

Considering $\kappa^{*}_{\max} > \kappa^{*}_{\min}$, the non-jackknife regions and jackknife limits (i.e., the boundaries of the non-jackknife regions) of the Long Trailer case can be derived per the following subcases:
\begin{enumerate}
\item{Subcase L-1:} if $\kappa^{}_{\max}>\kappa^{*}_{\max}$ and $\kappa^{}_{\min}<\kappa^{*}_{\min}$, there is only one non-jackknife region covering all hitch angle, i.e., hitch angle is always non-jackknifing. Thus, there is no non-jackknife region and no jackknife limit. This is the only subcase where all hitch angles are non-jackknifing.

\item{Subcase L-2:} if $\kappa^{}_{\max}>\kappa^{*}_{\max}$ and $\kappa^{}_{\min}\in[\kappa^{*}_{\min},\kappa^{*}_{\max}]$, there is only one non-jackknife region, which is $[\psi^{+}_{\kappa,\min},\psi^{-}_{\kappa,\min}]$. This non-jackknife region typically wraps around $\pm\pi$\,rad.

\item{Subcase L-3:} if  $\kappa^{}_{\max}\in[\kappa^{*}_{\min},\kappa^{*}_{\max}]$ and $\kappa^{}_{\min}<\kappa^{*}_{\min}$, there is only one non-jackknife region,  which is $[\psi^{-}_{\kappa,\max},\psi^{+}_{\kappa,\max}]$. This non-jackknife region typically wraps around $\pm\pi$\,rad.

\item{Subcase L-4:} if  $\kappa^{}_{\max}\le\kappa^{*}_{\max}$ and $\kappa^{}_{\min}\ge\kappa^{*}_{\min}$, there are two non-jackknife regions, Fig.\,\ref{fig:kappaV0_Long}. The first non-jackknife region is $[\psi^{-}_{\kappa,\max},\psi^{-}_{\kappa,\min}]$. The second non-jackknife region is $[\psi^{+}_{\kappa,\min},\psi^{+}_{\kappa,\max}]$. \textbf{This is the most common subcase in actual trailer tasks} among all the subcases including those of the Short and Medium Trailer case.  

\item{Subcase L-5} if  $\kappa^{}_{\max}<\kappa^{*}_{\min}$ or $\kappa^{}_{\min}>\kappa^{*}_{\max}$, there is no non-jackknife region and thus no jackknife limit. The trailer is always jackknifing.
\end{enumerate}

During calculation, some of the jackknife limits values can be outside the hitch angle range, $(-\pi,\pi]$\,rad. They should be converted into this range by moduloing by $2\pi$\,rad. If both boundaries are within the range and the lower limit is higher than the upper limit, it means the non-jackknife range expands across $\pm\pi$\,rad.

\section{Jackknife Limits} \label{jackknifeImpact}

According to Definition \ref{DefJK}, when hitch angle is in a jackknife region, the sign of hitch angle rate $\dot\psi$ stays constant. Since the two boundaries of a jackknife region are the jackknife limits of one or two non-jackknife regions, a constantly positive or negative ${\rm{sgn}}(\dot\psi)$ means that hitch angle always moves toward one jackknife limit and away from the other jackknife limit. If ${\rm{sgn}}(\dot\psi)=0$, hitch angle will be static. 

\subsection{Safe and Unsafe Jackknife Limits}

Based on the sign of hitch angle rate in the jackknife region adjacent to the jackknife limit, jackknife limits can be divided into two categories: safe and unsafe, which are defined as follows.

\begin{MyDef}
With given sideslip and sign of vehicle velocity, for a non-jackknife region:

A \textbf{Safe Jackknife Limit} is a jackknife limit where hitch angle would move back into the non-jackknife region if the hitch angle is in the adjacent jackknife region.

An \textbf{Unsafe Jackknife Limit} is a jackknife limit where hitch angle would move away from the non-jackknife region or stay constant if the hitch angle is in the adjacent jackknife region.
\label{DefJKType}
\end{MyDef}

In jackknife avoidance, a safe jackknife limit does not need to be considered, while an unsafe jackknife limit should be actively avoided. The general criteria for safe and unsafe jackknife limits is summarized as follows. 

\begin{MyTheo}
Let $\psi^{}_{JK1}$ be a jackknife limit and its corresponding vehicle curvature limit be $\kappa^{}_{{\rm{lim1}}}$. Let $\kappa^{}_{{\rm{lim2}}}$ be the other vehicle curvature limit.

\begin{enumerate}

\item If the non-jackknife region (if there exists one) is adjacent to and on the \textbf{positive side} of $\psi^{}_{JK1}$, $\psi^{}_{JK1}$ is a safe jackknife if $\dot\psi(\kappa^{}_{{\rm{lim2}}},\psi^{}_{JK1})>0$. Otherwise it is an unsafe jackknife limit.

\item If the non-jackknife region (if there exists one) is adjacent to and on the \textbf{negative side} of $\psi^{}_{JK1}$, $\psi^{}_{JK1}$ is a safe jackknife if $\dot\psi(\kappa^{}_{{\rm{lim2}}},\psi^{}_{JK1})<0$. Otherwise it is an unsafe jackknife limit.
\end{enumerate}
\label{TheoJKType}
\end{MyTheo}

The theorem is based upon the concept that a safe or unsafe jackknife limit depends on whether the hitch angle will increase or decrease such that hitch angle enters or leaves the non-jackknife region.  It is possible that the two conditions of the theorem occur at the same jackknife limit, resulting in the jackknife limit being safe for the non-jackknife region on one side and unsafe for the non-jackknife region on the other side. Detailed proof of the theorem is left to the reader.  

\subsection{Typical Safe and Unsafe Jackknife Limits}

Typically, a jackknife limit does not overlap with other jackknife limits and does not equal $\psi^{+}_{\infty}$ or $\psi^{-}_{\infty}$. It is much easier to determine whether a jackknife is safe or unsafe if it is a Typical Jackknife Limit. Typical Jackknife Limits are defined as follows.
\begin{MyDef} \label{DefTypicalJK}
For a jackknife limit, $\psi_{JK}$, and its corresponding achievable vehicle curvature limit, $\kappa^{}_{{\rm{lim1}}}$, $\psi_{JK}$ is a Typical Jackknife Limit if $\psi_{JK}\ne\psi^{+}_{\infty}$, $\psi_{JK}\ne\psi^{-}_{\infty}$, and $\alpha_1(\kappa^{}_{{\rm{lim1}}})\in(0,\pi)$.
\end{MyDef}

The criteria for safe and unsafe Typical Jackknife Limits is as follows.

\begin{MyTheo}
$\psi^{+}_{\kappa,\max}$ and $\psi^{+}_{\kappa,\min}$ are safe jackknife limits when $v<0$ and unsafe jackknife limit when $v>0$, if they are Typical Jackknife Limits. 

$\psi^{-}_{\kappa,\max}$ and $\psi^{-}_{\kappa,\min}$ are unsafe jackknife limits when $v<0$ and safe jackknife limit when $v>0$, if they are Typical Jackknife Limits. 
\label{TheoJKType2}
\end{MyTheo}

\begin{MyProof}

Consider a jackknife limit, $\psi^{}_{JK1}$, equal to $\psi^{+}_{\kappa,\max}$ or $\psi^{+}_{\kappa,\min}$. Denote the vehicle curvature limit corresponding to $\psi^{}_{JK1}$ as $\kappa^{}_{{\rm{lim1}}}$. According to (\ref{psiMaxIn}), (\ref{psiMinIn}), and (\ref{alpha1}), 
\begin{eqnarray}
\psi_{JK1}+\beta_T-\alpha_2(\kappa^{}_{{\rm{lim1}}})=\alpha_1(\kappa^{}_{{\rm{lim1}}})\in(0,\pi). \label{psiJKupperRelation}
\end{eqnarray}

According to (\ref{modelr4}), $\dot\psi$ can be expressed as a function of $\kappa$ and $\psi$,
\begin{eqnarray}
\dot\psi(\kappa,\psi) &=&-v\alpha_3[\cos(\psi+\beta_T-\alpha_2(\kappa)) \nonumber\\
& &-\cos(\alpha_1(\kappa))] \label{dotPsiFunc}
\end{eqnarray}
\noindent where $\alpha_1$ is shown in (\ref{alpha1}), $\alpha_2$ is shown in (\ref{alpha2}), and $\alpha_3$ is
\begin{eqnarray}
\alpha_3 =\frac{\sqrt{L^2_1\kappa^{2}_{}+1-2L_1\kappa^{}_{}\sin\beta_R}}{L_2\cos\beta_T}. \label{alpha3}
\end{eqnarray}

For a given $\kappa$, it can be derived from (\ref{dotPsiFunc}) that
\begin{eqnarray}
\frac{\partial\dot\psi}{\partial\psi}=v\alpha_3\sin(\psi+\beta_T-\alpha_2(\kappa)). \label{dDotPsi_dPsi}
\end{eqnarray}

Substitute (\ref{psiJKupperRelation}) in to (\ref{dDotPsi_dPsi}) at $\psi=\psi^{}_{JK1}$ and $\kappa=\kappa^{}_{{\rm{lim1}}}$,
\begin{eqnarray}
&&\left.\frac{\partial\dot\psi(\kappa^{}_{{\rm{lim1}}},\psi)}{\partial\psi}\right|_{\psi=\psi_{JK1}} \nonumber\\
&&=v\alpha_3\sin(\psi_{JK1}+\beta_T-\alpha_2(\kappa^{}_{{\rm{lim1}}}))  \nonumber \\
&&= v\alpha_3\sin(\alpha_1(\kappa^{}_{{\rm{lim1}}}))
\label{dDotPsi_dPsi1}
\end{eqnarray}
\noindent where, according to (\ref{alpha3}) and the nonoverlapping conditions, it can be found that $\alpha_3>0$ and $\alpha_1(\kappa^{}_{})\in(0,\pi)$ (i.e., $\sin(\alpha_1(\kappa^{}_{{\rm{lim1}}}))>0$). Thus, (\ref{dDotPsi_dPsi1}) indicates ${\rm{sgn}}(\partial\dot\psi(\kappa^{}_{{\rm{lim1}}},\psi)/\partial\psi)$ $={\rm{sgn}}(v)$ at $\psi=\psi^{}_{JK1}$.

Let $\psi_{JK1L}$ be the hitch angle point adjacent to and on the left side of $\psi_{JK1}$ (i.e. less than $\psi_{JK1}$).

Assume $v>0$. Then ${\rm{sgn}}(\partial\dot\psi(\kappa^{}_{{\rm{lim1}}},\psi)/\partial\psi)>0$. Since ${\rm{sgn}}(\dot\psi(\kappa^{}_{{\rm{lim1}}},\psi_{JK1}))=0$, there is ${\rm{sgn}}(\dot\psi(\kappa^{}_{{\rm{lim1}}},\psi_{JK1L}))<0$. Assume there is a jackknife region adjacent to $\psi_{JK1}$ and on its left side. $\psi_{JK1L}$ is in this jackknife region. Since ${\rm{sgn}}(\dot\psi(\kappa^{}_{{\rm{lim1}}},\psi_{JK1L}))<0$, all hitch angle in this jackknife region would result in negative $\dot\psi$ regardless of the achievable vehicle curvature applied. Thus, hitch angle would move farther away from $\psi_{JK1}$ once it is in the jackknife region. This makes $\psi_{JK1}$ an unsafe jackknife limit. It can be proven in a similar way that $\psi_{JK1}$ is also an unsafe jackknife limit if the jackknife region is on its right side.

Similarly, it can be found that when $v<0$, $\psi_{JK1}$ is a safe jackknife limit. Thus, $\psi^{+}_{\kappa,\max}$ and $\psi^{+}_{\kappa,\min}$ are safe jackknife limits when $v<0$ and unsafe jackknife limit when $v>0$, if they are Typical Jackknife Limits.

Similarly, it can be proven that $\psi^{-}_{\kappa,\max}$ and $\psi^{-}_{\kappa,\min}$ are unsafe jackknife limits when $v<0$ and safe jackknife limit when $v>0$, if they are Typical Jackknife Limits.
\qedsymbol
\end{MyProof}

\section{Results and Discussion} \label{section_test}

Simulation and experimental results verify the jackknife analysis discussed above. Simulations demonstrate the jackknifing process and verify the jackknife limits illustrated earlier. Experimental results validate jackknife limit calculations.

\subsection{Simulation Results for Jackknifing Behaviors} \label{SEC_Simulation1}

This subsection verifies jackknifing analysis using simulations where hitch length is 1.23\,m, steering ratio is 17.6, and vehicle wheel base is 3\,m.

\begin{figure}
\centerline{\epsfig{figure=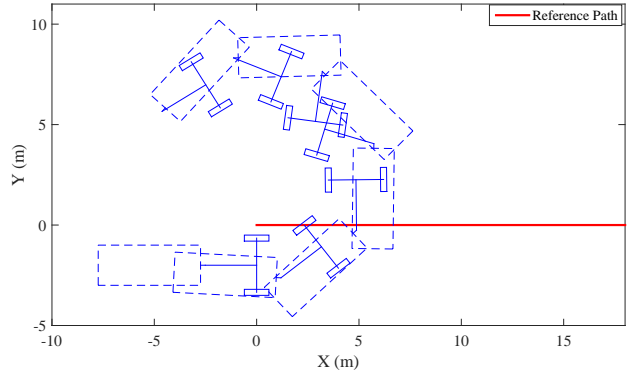,height=5cm}}
\vspace{-0.1 cm}
\caption{An example of Long Trailer jackknifing process during trailer backing, with 2m lateral error.}
\label{fig:LongSimu1}
\end{figure}

\begin{figure}
\centerline{\epsfig{figure=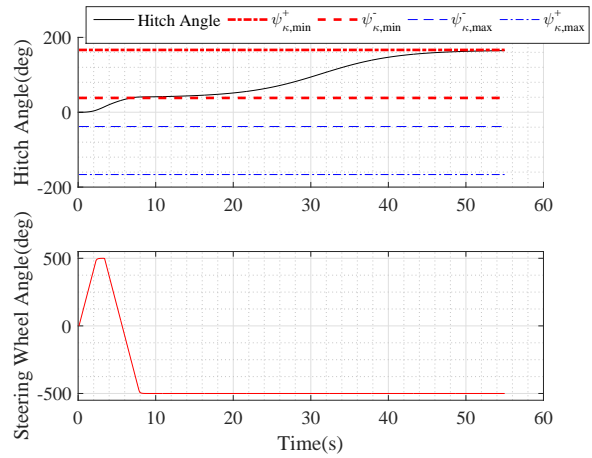,height=6cm}}
\vspace{-0.1 cm}
\caption{Hitch angle and steering wheel angle of the Long Trailer jackknifing process shown in Fig.\,\ref{fig:LongSimu1}.}
\label{fig:LongSimu2}
\end{figure}

A typical jackknifing scenario is shown in Fig.\,\ref{fig:LongSimu1} and Fig.\,\ref{fig:LongSimu2} for a Long Trailer case with trailer tongue length 2.51\,m, steering wheel angle limit $\pm500^\circ$, and zero sideslip. The vehicle and trailer initially are parallel to the path and 2\,m from the path. The system is controlled by the backing control law deveploped in \cite{Pradalier2007} and revised in \cite{Leng2016} without its jackknife recovery algorithm. Since the initial lateral tracking error (2 meters) is high, the controller tries to compensate it with an extreme steering angle, Fig.\,\ref{fig:LongSimu2}. After about 3 s, the controller decides that the hitch angle is too large and commands the steering wheel to move towards -500$^\circ$ so as to reduce hitch angle magnitude. Though the controller tries to straighten the vehicle at around 4 s, the hitch angle exceeds the jackknife limit at 8 s due to limited steering wheel rate. Per Sec.\ref{jackknifeImpact}, hitch angle becomes jackknifing and approaches the safe jackknife limit (i.e. $\psi^+_{\kappa,\min}$) on the other side of the jackknife region. 

\begin{figure}
\centerline{\epsfig{figure=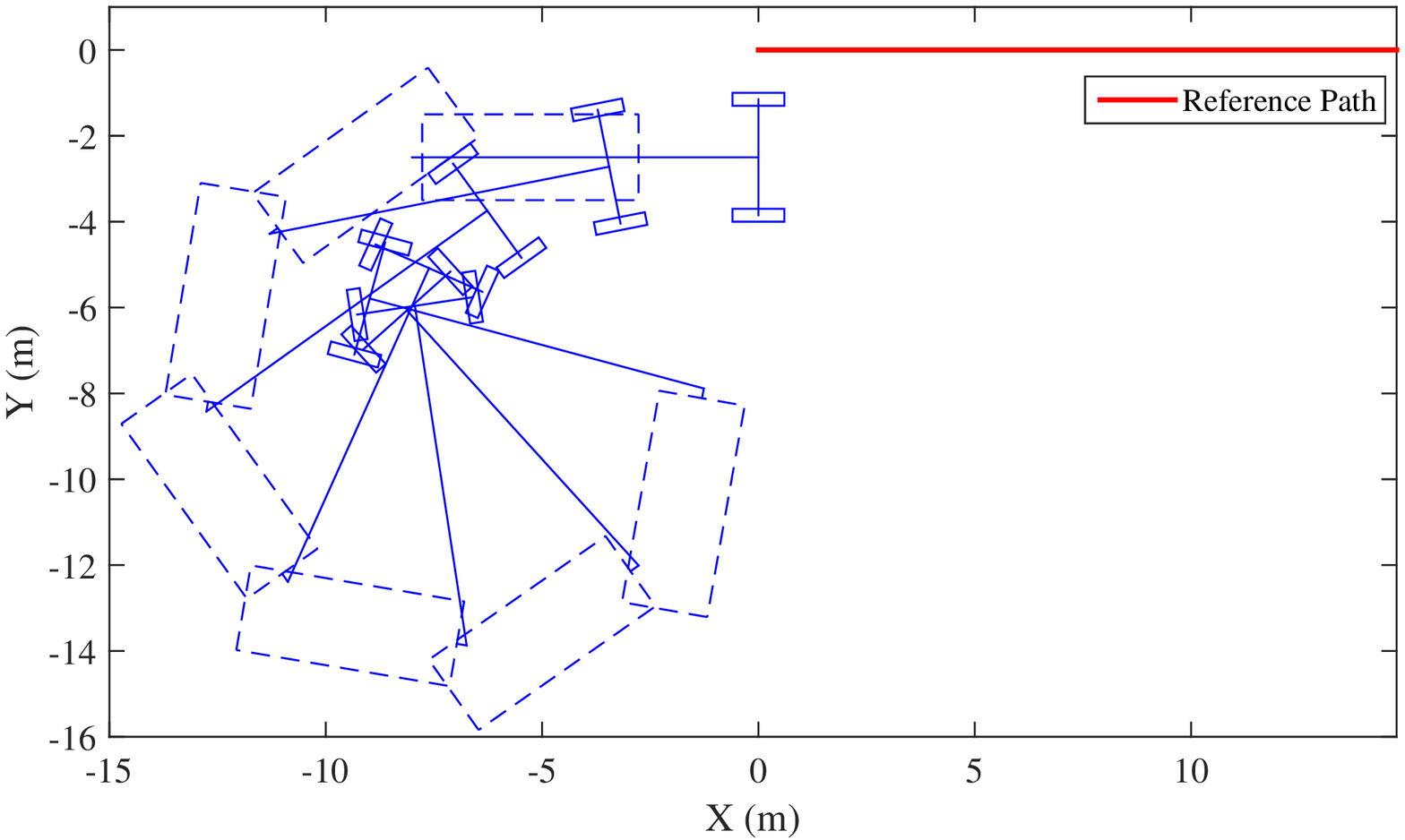,height=5cm}}
\caption{An example of Long Trailer movement without jackknife limit, during trailer backing.}
\label{fig:LongTrailerSimuRun1_1}
\end{figure}

\begin{figure}
\centerline{\epsfig{figure=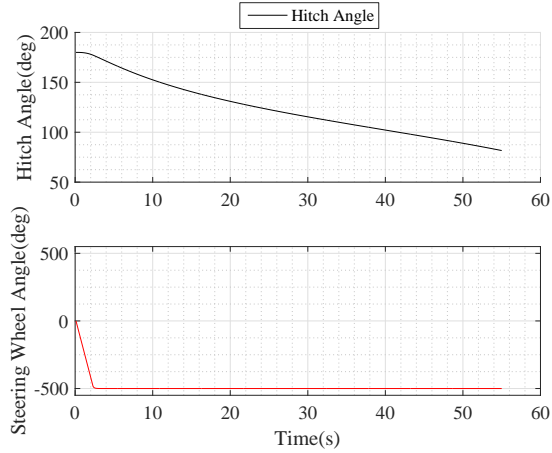,height=6cm}}
\caption{Hitch angle and steering wheel angle of the movement shown in Fig.\,\ref{fig:LongTrailerSimuRun1_1}.}
\label{fig:LongTrailerSimuRun1_2}
\end{figure}

The second simulation is shown in Fig.\,\ref{fig:LongTrailerSimuRun1_1} and Fig.\,\ref{fig:LongTrailerSimuRun1_2}. The trailer starts parallel to the reference path at $x=0$ with $\psi=180^\circ$. This illustrates a scenario predicted in Case L-1 in Sec.\,\ref{longTrailer} where the system is always non-jackknifing and there is no jackknife limit. Physical hitch angle limits defined by mechanical structure are ignored in this simulation in order to better demonstrate Case L-1. A difference between the Long Trailer and the Short and Medium Trailer is that the vehicle curvature needed to maintain a hitch angle is always limited for a Long Trailer, Fig.\,\ref{fig:kappaV0_Long}, but may be infinite for the Short and Medium Trailers, Fig.\,\ref{fig:kappaV0_Short} and  Fig.\,\ref{fig:kappaV0_Medium}, respectively. Thus, a Long Trailer system may not have jackknife limits and jackknife regions if the achievable vehicle curvature range is large enough. In this simulation, the parameter settings are as follows: $L_2=8$\,m, steering wheel angle limit is $\pm500^\circ$, initial lateral error is 2.5\,m, and zero sideslip. In this scenario, the trailer starts at $X=0$. Initial hitch angle is 180$^\circ$ to highlight an extreme situation. The controller gradually moves the hitch angle towards zero radians, although Fig.\,\ref{fig:LongTrailerSimuRun1_1} is truncated at 55 s for clarity.

\begin{figure}
\centerline{\epsfig{figure=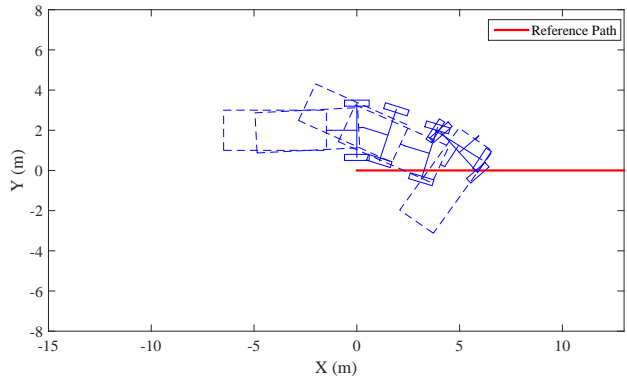,height=5cm}}
\caption{An example of Medium Trailer jackknifing process during trailer backing, with 2.5\,m initial lateral error.}
\label{fig:MediumTrailerSimuRun1_1}
\end{figure}

\begin{figure}
\centerline{\epsfig{figure=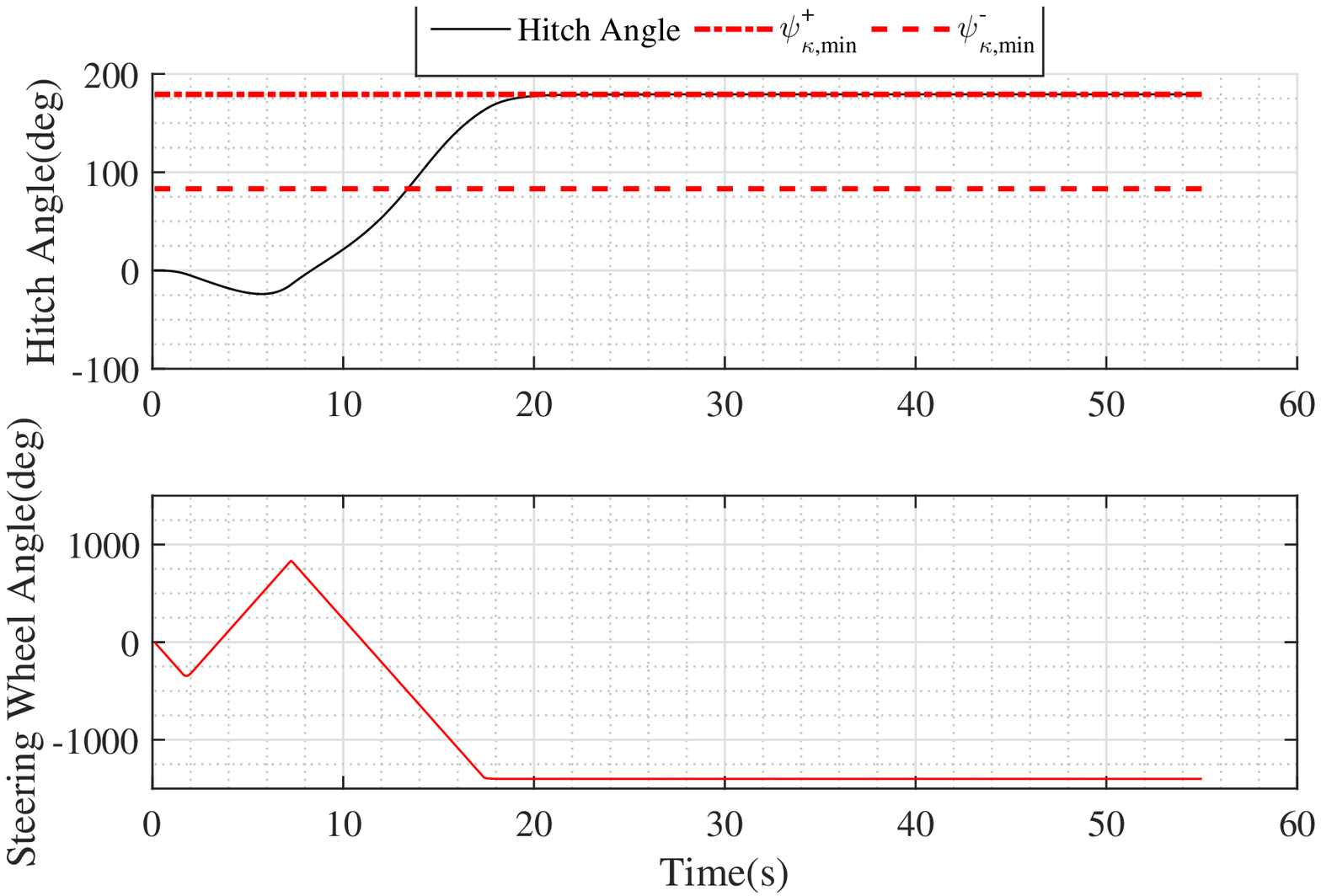,height=6cm}}
\caption{Hitch angle and steering wheel angle of the movement shown in Fig.\,\ref{fig:MediumTrailerSimuRun1_1}.}
\label{fig:MediumTrailerSimuRun1_2}
\end{figure}

The third simulation demonstrates the Medium Trailer case, i.e. the Subcase M-8 discussed in Sec.\,\ref{mediumTrailer}, Fig.\,\ref{fig:MediumTrailerSimuRun1_1} and Fig.\,\ref{fig:MediumTrailerSimuRun1_2}. The parameters are $L_2=1.25$\,m, $\pm1400^\circ$ steering wheel angle limit (which is not realistic, but suitable for demonstration purposes ), 2.5\,m intial lateral error, and $\beta_R=\beta_T=30^\circ$. In this case, there are only two jackknife limits, i.e., $\psi^{+}_{\kappa,\min}$ and $\psi^{-}_{\kappa,\min}$, Fig.\,\ref{fig:MediumTrailerSimuRun1_2}. The region between them (i.e., the hitch angle range between 83$^\circ$ and 179$^\circ$) is the jackknife region, while other hitch angles are all non-jackknifing. $\psi^{-}_{\kappa,\min}$ is an unsafe jackknife limit and $\psi^{+}_{\kappa,\min}$ is a safe jackknife limit. Thus, when hitch angle exceeds $\psi^{-}_{\kappa,\min}$ and enters the jackknife region, it keeps increasing regardless of what achievable vehicle curvature is used. Per Fig.\,\ref{fig:MediumTrailerSimuRun1_2}, hitch angle eventually converges to the safe jackknife limit $\psi^{+}_{\kappa,\min}$, at which point the system can freely move hitch angle in the non-jackknife region. Note that the results shown in Fig.\,\ref{fig:MediumTrailerSimuRun1_1} are truncated at around 20 seconds for clarity. 

\begin{figure}
\centerline{\epsfig{figure=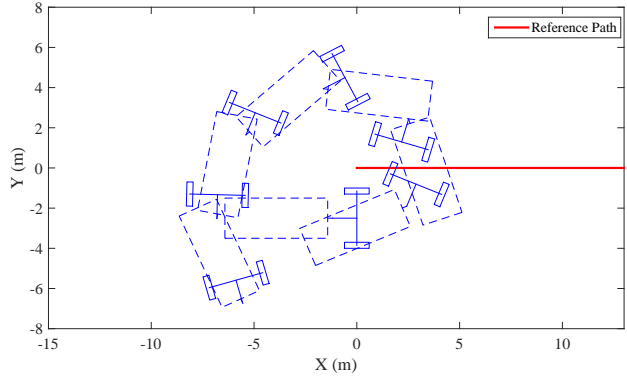,height=5cm}}
\caption{An example of Medium Trailer jackknifing process during trailer backing, with varying sideslip and 2.5\,m initial lateral error.}
\label{fig:SideslipTrailerSimuRun1_1}
\end{figure}

\begin{figure}
\centerline{\epsfig{figure=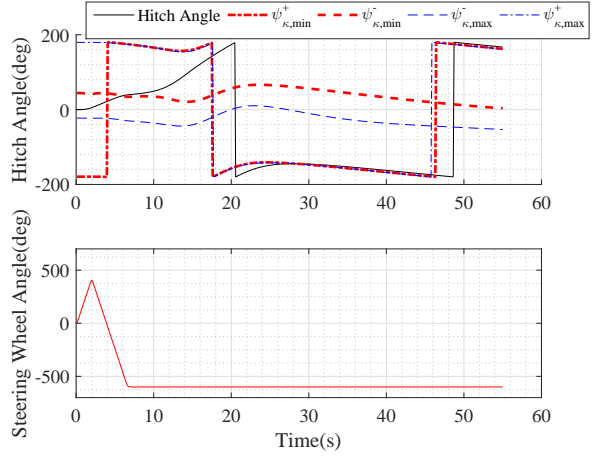,height=6cm}}
\caption{Hitch angle and steering wheel angle of the movement shown in Fig.\,\ref{fig:SideslipTrailerSimuRun1_1}.}
\label{fig:SideslipTrailerSimuRun1_2}
\end{figure}

The fourth simulation demonstrates sideslip varying relative to terrain orienation, Fig.\,\ref{fig:SideslipTrailerSimuRun1_1} and Fig.\,\ref{fig:SideslipTrailerSimuRun1_2}. The parameters are  $L_2=1.25$\,m, steering wheel angle limit is $\pm600^\circ$, and intial lateral error is 2.5\,m. Sideslips are:
\begin{eqnarray}
\beta_F &=& \pi\cos(\theta^{}_V+\phi)/6\\
\beta_R &=& \pi\cos(\theta^{}_V)/6\\
\beta_T &=& \pi\cos(\theta^{}_T)/6,
\end{eqnarray}
\noindent where $\theta^{}_V$ is vehicle heading, $\theta^{}_T$ is trailer heading, and $\phi$ is steering angle. Note that hitch angle and jackknife limits are converted into the range of $(-180^\circ,180^\circ]$. Thus, when hitch angle or a jackknife limit rise higher than 180$^\circ$, it would jump to near -180$^\circ$. In this scenario, the trailer starts at $X=0$ with zero hitch angle. Hitch angle exceeds the jackknife limit at ~6 s because of the decrease of the jackknife limit $\psi^{-}_{\kappa,\min}$  due to sideslip variation. If the jackknife limit $\psi^{-}_{\kappa,\min}$ did not decrease due to sideslip, the system would not jackknife at this point. Hitch angle then moves towards the neighboring safe jackknife limit, $\psi^{+}_{\kappa,\max}$. Then between 20 s and 50 s, hitch angle crosses the non-jackknife region between the two safe jackknife limits, i.e. $\psi^{+}_{\kappa,\min}$ and $\psi^{+}_{\kappa,\max}$ due to sideslip variation. It moves into the other jackknife region and approaches the neighboring safe jackknife limit, i.e. $\psi^{+}_{\kappa,\min}$.

\begin{figure}
\centerline{\epsfig{figure=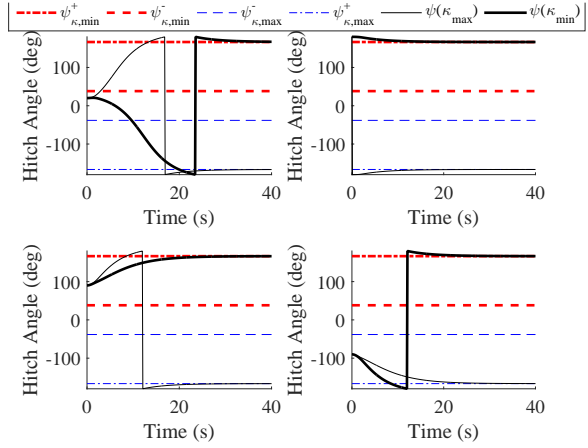,height=6cm}}
\vspace{-0.3 cm}
\caption{Simulations showing hitch angle with system starting in jackknife and non-jackknife regions with constant steering angle.}
\label{fig:HitchAngleSimu}
\end{figure}

Fig.\,\ref{fig:HitchAngleSimu} shows how hitch angle develops when it starts in various jackknife and non-jackknife regions under constant steering wheel angle during trailer backing. $\psi(\kappa^{}_{\max})$ and $\psi(\kappa^{}_{\min})$ are the hitch angle trajectories resulting from the maximum and the minimum vehicle curvature ($\kappa^{}_{\max}$ and $\kappa^{}_{\min}$), respectively. If vehicle curvature is fixed at any achievable curvature other than the two limits, then the hitch angle trajectory would be in between the two trajectories shown in each subplot with $\psi(\kappa^{}_{\max})$ as the upper bound and $\psi(\kappa^{}_{\min})$ as the lower bound. The four jackknife limits ($\psi^{+}_{\kappa,\max}$, $\psi^{-}_{\kappa,\max}$, $\psi^{+}_{\kappa,\min}$, and $\psi^{-}_{\kappa,\min}$) are also shown in each subplot. In this case, $\psi^{+}_{\kappa,\max}$ and $\psi^{+}_{\kappa,\min}$ are safe jackknife limits, while $\psi^{-}_{\kappa,\max}$ and $\psi^{-}_{\kappa,\min}$ are unsafe jackknife limits. 

The upper left subplot shows hitch angle trajectories based upon maximum ($\psi(\kappa^{}_{\max})$) and minimum ($\psi(\kappa^{}_{\min})$) vechicle curvature when hitch angle starts in the non-jackknife region with two unsafe jackknife limits (i.e., $\psi^{-}_{\kappa,\max}$ and $\psi^{-}_{\kappa,\min}$ in this case). Hitch angle soon exceeds one of the unsafe jackknife limits, moves across a jackknife region, and finally enters the non-jackknife region with two safe jackknife limits. Once hitch angle exceeds an unsafe jackknife limit, it cannot move back unless the vehicle and trailer changes its movement direction. Notice that the hitch angle trajectory (i.e., $\psi(\kappa^{}_{\max})$) resulting from the maximum vehicle curvature eventually converges at the safe jackknife limit (i.e., $\psi^{+}_{\kappa,\max}$) resulting from the maximum vehicle curvature. Similarly, the hitch angle trajectory (i.e., $\psi(\kappa^{}_{\min})$) resulting from the minimum vehicle curvature eventually converges at the safe jackknife limit (i.e., $\psi^{+}_{\kappa,\min}$) resulting from the minimum vehicle curvature. Similar convergence processes appear in all other subplots.

The upper right subplot shows the hitch angle trajectories starting in the non-jackknife region with two safe jackknife limits. Recall that a hitch angle trajectory under any achievable vehicle curvature would be between $\psi(\kappa^{}_{\max})$ and $\psi(\kappa^{}_{\min})$. Per Sec.\, \ref{jackknifeImpact}, hitch angle cannot cross a safe jackknife limit from a non-jackknife region with any achievable vehicle curvature.

The lower left and lower right subplots show hitch angle trajectories starting in the jackknife regions. Note that hitch angles starting in the jackknife region always move toward the non-jackknife region with two safe jackknife limits, regardless of vehicle curvature. 

It can also be noticed that when hitch angle is in a non-jackknife region, hitch angle can be manipulated to increase and decrease as desired by commanding vehicle curvature within its achievable range. Typically, when hitch angle is in a jackknife region, hitch angle always increases or decreases regardless of vehicle curvature within its achievable limits, until hitch angle reaches one of the two safe jackknife limits. As proven in Theorem \ref{OverlappingTheo}, hitch angle stays constant in a jackknife region if and only if the jackknife region consists of the overlapping jackknife limits in Case O-1. 

\begin{figure}
\centerline{\epsfig{figure=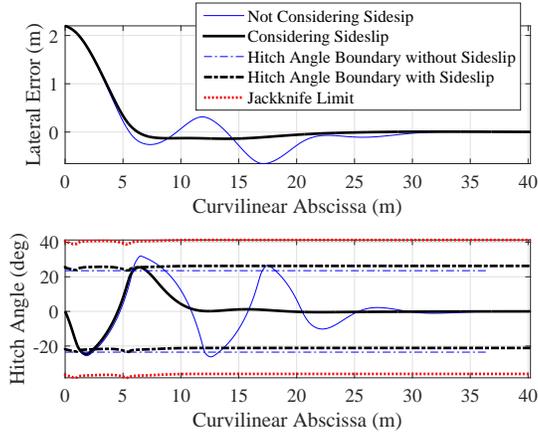,height=6cm}}
\vspace{-0.1 cm}
\caption{Trailer backing control results considering the effect of sideslip on jackknifing limits compared to using traditional fixed jackknife limits with a simulated sideslope.}
\label{fig:figControl}
\end{figure}

Existing controllers can benefit from the jackknife analysis in this paper. As an example, the Pradalier \cite{PradalierJFR} controller modified per \cite{Leng2016} is revised to consider side-slip. Hitch angle boundaries that are used to trigger jackknife corrections are modified to consider the effect of sideslip.  A 15 degree hitch angle boundary is implemented relative to jackknife limits such that the controller cannot command hitch angles outside of the nonjackknife zone; pulling forward to straighten the hitch angle is not used here.  Trailer backing performance is compared using the slip-based jackknife limits in this paper compared to using jackknife limits derived without slip.  Long trailer backing given an initial 2.2 m lateral error is performed along a straight path with a sideslope simulated by 5 degrees sideslip perpendicular to the path. When sideslip is not considered, hitch angle gets too close to the jackknife limit, Fig.\,\ref{fig:figControl}. resulting in reduced controllability of hitch angle and subsequent oscillations and longer convergence.  When the effect of sideslip is considered, jackknife limits are adjusted and steering angles are limited appropriately to prevent the system from approaching the jackknife limit.  As a result, the controller does not produce oscillations and quickly drives lateral error to zero.

\subsection{The Impact of Sideslip}

\begin{figure}
\centerline{\epsfig{figure=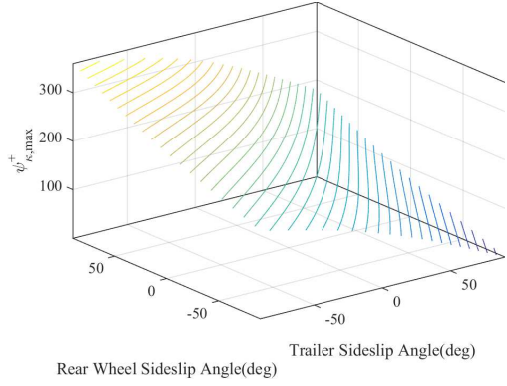,height=5cm}}
\vspace{-0.1 cm}
\caption{Jackknife limit $\psi^{+}_{\kappa,\max}$ with respect to rear wheel sideslip angle $\beta_R$ and trailer sideslip angle $\beta_T$.}
\label{fig:PsiUpperPlusContour}
\end{figure}

Recall (\ref{psiMaxIn}) to (\ref{alpha2}). The four jackknife limits are influenced by rear wheel sideslip angle and trailer sideslip angle, i.e., $\beta_T$ and $\beta_R$. Recall that $\beta_T$ and $\beta_R$ are within $[-\pi/2,\pi/2]$.  Fig.\,\ref{fig:PsiUpperPlusContour} shows the contour plot of one of the jackknife limits, $\psi_{\uparrow,+}$, with respect to $\beta_T$ and $\beta_R$. It can be observed that the contour is nearly flat, meaning that the jackknife limit essentially increases and decreases linearly with respect to sideslip.  

\subsection{Experimental Results}

Two physical experiments were conducted to validate derivations. The parameters of the vehicle-trailer system used for experiments are the same as those shown in Sec.\,\ref{SEC_Simulation1}.

\begin{figure}
\centerline{\epsfig{figure=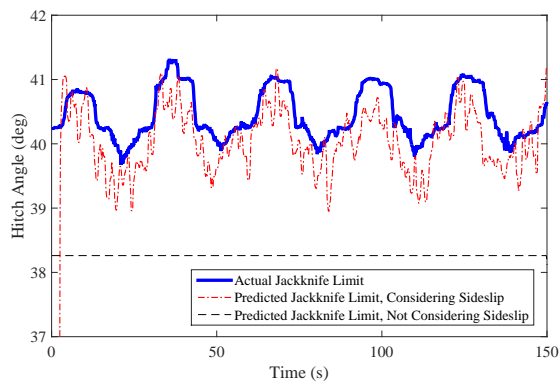,height=5cm}}
\vspace{-0.1 cm}
\caption{Predicted jackknife limits and actual jackknife limit for forward trailer movement.}
\label{fig:ActualJKL_Forward}
\end{figure}

The first experiment is designed to validate the safe jackknife limit. This is done by turning the steering wheel to its physical limit, 500 degrees, and driving the vehicle-trailer system forward. The corresponding steering wheel angle can be calculated using vehicle wheel base, $L$, and steering ratio 17.6 provided earlier. During this test, the vehicle and trailer system is driving on a large parking lot with a moderate slope, which changes orientation to the vehicle as it traverses the trajectory. This causes periodic variation in sideslip and the actual jackknife limit. The actual and predicted jackknife limits are shown in Fig.\,\ref{fig:ActualJKL_Forward}. The actual jackknife limit data shown are from the hitch angle data measured by the hitch angle sensor. As analyzed in Sec.\,\ref{jackknifeImpact} and simulated earlier, if hitch angle starts in the non-jackknife region with two safe jackknife limits or in either jackknife region, and the steering wheel is set to one of its limits, then the hitch angle will always converge to the safe jackknife limit resulting from the steering wheel limit used. Since the jackknife limit changed slowly during this test, the difference between the hitch angle and the jackknife limit can be ignored. Thus, we can consider the hitch angle in this test as the actual jackknife limit.

There are two predicted jackknife limits  in Fig.\,\ref{fig:ActualJKL_Forward}. The first one is predicted with sideslip, which is calculated from the vehicle yaw rate measured by an IMU and vehicle speed measured by GPS. It partially considers sideslip, as it ignores the sideslip in the calculation (\ref{psiMaxIn}-\ref{psiMinOut}) but considered the impact of sideslip on vehicle curvature. Due to sensor noise, this prediction is filtered with a two-sided averaging filter with a window width of 9. This prediction is quite close to the actual jackknife limit. \
The second prediction is calculated from vehicle curvature in the basic vehicle model (\ref{kappaV}), but sideslip is ignored, which highlights the traditional jacknife limit estimate. The traditional estimate is about $2^\circ$ lower than the nominal value of the jackknife limit including sideslip, which is attributed to sideslip from the circular trajectory and vehicle model inaccuracy.

\begin{figure}
\centerline{\epsfig{figure=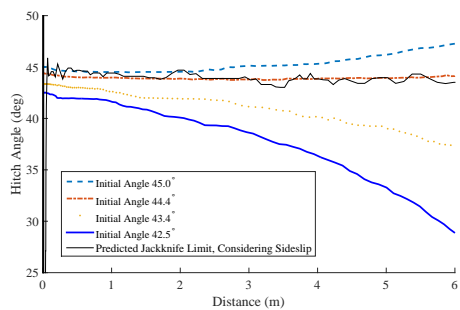,height=4cm}}
\vspace{-0.1 cm}
\caption{Predicted jackknife limit and hitch angle trajectories with various intial values for trailer backing.}
\label{fig:ActualJKL_Backing}
\end{figure}

The second experiment validates the unsafe jackknife limit by backing the vehicle-trailer system with various initial hitch angles down a sloped surface. Four initial hitch angles are used to estimate the unsafe jackknife limit, Fig.\,\ref{fig:ActualJKL_Backing}. Results suggest that the actual unsafe jackknife limit is approximately 44.4$^\circ$. However, one can notice the slight changes in the ``$44.4^\circ$" hitch angle trajectory, which are attributed to variation of sideslope and slight uneveness in the ground as the system backs along the path. The actual unsafe jackknife limit is very close to the predicted unsafe jackknife limit, around 44$^\circ$. This jackknife limit is predicted using the data from the experiment with the 45$^\circ$ initial hitch angle. The jackknife limits predicted using the data from the other initial conditions are essentially the same but are omitted for clarity due to noise in the estimates. The predicted jackknife limit is calculated similar to the forward test, i.e., using the vehicle curvature calculated from yaw rate and vehicle speed while partially taking sideslip into consideration, and is again filtered with a two-sided averaging filter with a window width of 9.

Finally, per Sec.\,\ref{jackknife}, the jackknife limits in backing and forward should be the same when the steering wheel is set to the same angle. However, this is not the case in experiments. The actual jackknife limit going forward is around 40.5$^\circ$, Fig.\,\ref{fig:ActualJKL_Forward}, whereas the actual jackknife limit in backing is around 44.4$^\circ$, Fig.\,\ref{fig:ActualJKL_Backing}. We believe this difference is due to mechanics of the vehicle (e.g., Ackerman steering, wheel camber, etc.) that causes the vehicle curvature to be different in backing and forward operations even though the steering wheel angles are the same. Future work could focus on revisiting the analysis using an improved vehicle model to better capture the subtlety of the steering system.

\section{Conclusions and Future Work} \label{Conclusion}

Jackknifing is analyzed in detail with sideslip taken into consideration. For the analysis, vehicle-trailer systems are divided into three categories based on the ratio between hitch length and trailer tongue length. Jackknife limits are derived as well as the jackknife and non-jackknife hitch angle regions. Jackknife limits are divided into two categories, i.e., safe and unsafe. Unsafe jackknife limits should be avoided, while safe jackknife limits would not cause problems. Simulation results verify analytical predictions regarding the effect of slip on jackknife limits and demonstrate the benefit of considering the effect of slip on jackknife limits during trailer operation. Jackknifing behavior near safe and unsafe limits is confirmed.  Experimental results demonstrate the ability to estimate jackknife limits using onboard sensors. This paper provides insight about hitch angles that lead to jackknifing, which could be used to derive improved trailer backing controllers in future work. Such information should help controllers better avoid jackknfing while providing tracking.  The effect of dynamics on slip and jackknifing could also be considered in future work to provide improved performance.

\section*{ACKNOWLEDGMENTS}

The authors would like to thank Kairos Autonomi for their generous support.

%
% ...



\section*{Nomenclature} \label{Nomenclature}
The following nomenclature is used in this paper:
\begin{itemize}
\item $B$, Front axle center
\item $C$, Rear axle center
\item $H$, Hitch point
\item $Q$, Trailer axle center
\item $(x,y)$, Vehicle coordinate, the coordinate of $C$
\item $(x_T,y_T)$, Trailer coordinate, the coordinate of $Q$
\item $v$, Vehicle speed
\item $\phi$, Steering angle 
\item $\theta_V$, Vehicle heading
\item $\theta_T$, Trailer heading
\item $\psi$, Hitch angle
\item $L$, Vehicle wheel base 
\item $L_1$, Hitch length 
\item $L_2$, Trailer tongue length 
\item $\beta_F$, Front wheel sideslip 
\item $\beta_R$, Rear wheel sideslip 
\item $\beta_T$, Trailer wheel sideslip 
\item $\kappa$, Vehicle movement curvature 
\item $\kappa^{}_{\min}$,$\kappa^{}_{\max}$, Min/Max achievable vehicle curvature. 
\item $\kappa^*_{}(\psi)$, Critical Vehicle Curvature  
\item $\kappa^{*}_{\max}$, $\kappa^{*}_{\min}$, Local min/max of $\kappa^*_{}$ 
\item $\psi^{+}_{\kappa,\max}$, $\psi^{-}_{\kappa,\max}$, Jackknife limits resulted from $\kappa^{}_{\max}$ 
\item $\psi^{+}_{\kappa,\min}$, $\psi^{-}_{\kappa,\min}$, Jackknife limits resulted from $\kappa^{}_{\min}$ 
\item $\psi^{+}_{\infty}$, $\psi^{-}_{\infty}$, Uncontrollable Hitch Angles 
\end{itemize}

\section{References}

\bibliographystyle{tfnlm}
\bibliography{PaperRefs}

\end{document}